\documentclass{article}
\usepackage{spconf,amsmath,graphicx}

\usepackage{amsfonts}       
\usepackage{nicefrac}       
\usepackage{microtype}      
\usepackage{lipsum}
\usepackage{bm}
\usepackage{float}

\DeclareMathOperator*{\argmax}{arg\,max}

\usepackage{mathtools}

\usepackage{empheq}

\title{A Treatise on FST Lattice based MMI training}
\name{Adnan Haider, Tim Ng, Zhen Huang, Xingyu Na and  Antti Veikko Rosti}

 \address{Apple \\
\{adnan\_haider, kwongtim\_ng,  zhen\_huang,  na\_xingyu, arosti\}@apple.com   }

\begin{document}
%
\maketitle
\begin{abstract}
Maximum mutual information (MMI) has become one of the two  de facto methods for sequence-level training of speech recognition acoustic models. This paper aims to isolate, identify and bring forward  the implicit modelling decisions induced by the  design implementation of standard  finite state transducer (FST) lattice based  MMI training framework.  The paper  particularly investigates the necessity to maintain a pre-selected  numerator alignment and raises the importance of determinizing FST denominator lattices on the fly. The efficacy of employing on the fly FST lattice determinization is  mathematically shown to guarantee  discrimination at the hypothesis level and is empirically shown through training deep CNN models on a 18K  hours Mandarin dataset  and on a  2.8K hours English dataset. On assistant and dictation tasks, the approach achieves between   2.3-4.6\%  relative WER reduction (WERR) over the standard FST lattice based approach.

\end{abstract}
\section{Introduction}
\label{sec:intro}
Maximum mutual information (MMI)  \cite{bahl1986maximum} is one of the two main approaches that has been experimentally shown to be the most effective in training state of  the art acoustic models comprising of   hidden Markov models (HMMs)\cite{Baum1967}  embedded with deep neural networks (DNNs) starting from a good initialisation \cite{Xiong:2016a,Saon:2016a}.  When context-dependent phone HMMs serve as the basic model to construct sentence level HMMs, it is computationally infeasible to train these hybrid  models exactly with the MMI loss. In practice, models are trained within a computationally tractable lattice based framework \cite{MMI2,Povey2005}.

This paper brings forward the  implicit modifications  introduced to the MMI objective when implemented within  the FST lattice based framework. In addition, this paper makes the following novel contributions:
\begin{enumerate}
    \item This work presents a lower-bound proof that shows how dynamically chosing the Viterbi state alignment of each competing hypothesis in the denominator lattice leads to improvement in the original MMI objective. The efficacy of the proposed approach is shown through training deep CNN models on a 18K hours  Mandarin dataset  and on a  2.8K  hours English dataset. 
   \item This work mathematically shows the effectiveness of using a fixed numerator alignment during MMI loss computation. The efficacy of using a fixed alignment against dynamically sampling the alignment for the MMI numerator loss computation is empirically verified on the 2.8K English dataset.
   \end{enumerate}
The rest of the paper is organised as follows: Sec. \ref{sec:MMI} reviews MMI training. Sec. \ref{sec:fst} describes the FST lattice based MMI training framework. Sec. \ref{sec:latdet} presents the proposed approach with the proof given in the sec. \ref{sec:theroem}. Sec. \ref{sec:setup} presents  the experimental setup  followed by experimental results and conclusion.
 \section{MMI Training}
\label{sec:MMI}

To facilitate understanding of the mathematical analysis undertaken in this work, let  $(\mathbf{O},\mathbf{W}^{\text{ref}})$  denote a sample where $\mathbf{W}^{\text{ref}}$ is the reference word sequence associated with the observation feature sequence $\mathbf{O}$. In the context of lattice based sequence training, a 3 state  HMM topology \cite{htk,kaldi}  equipped with   emitting states serve  as the underlying model for individual phones, the basic unit of the AM.  An important advantage of using this modular form of HMM topology is that larger HMMs can be constructed by the concatenation or composition in the case of FSTs \cite{mohri2002weighted} of these basic models. This property allows   sentence models to be constructed providing the flexibility to integrate LM scores as transition probabilities between states matching the start and end of words. In this work, the set of state alignments, indexed by $i$, generated from the sentence HMM of the $j$th hypothesis $\mathbf{W}^j$   w.r.t to   $\mathbf{O}$ of $T$ frames will be denoted as  $\{\bm{y}^{i,j}_{1:T}\}_i$. For a given  $\mathbf{O}$, the MMI loss is:
 \begin{align}
\label{eq:MMIloss}
\mathcal{L}_{{MMI}}(\bm{\theta}) = \mathcal{G}(\bm{\theta})  -  \mathcal{F}(\bm{\theta}) \text{ where}
\end {align}
\begin{align}
    \mathcal{G}(\bm{\theta}) &= \text{log} \sum_{j} P_{\bm{\theta}}(\mathbf{O}|\mathbf{W}^j)P(\mathbf{W}^j) \notag \\
    &=\text{log} \sum_{j} \sum_i P_{\bm{\theta}}(\bm{y}^{i,j}_{1:T},\mathbf{O}|\mathbf{W}^j)P(\mathbf{W}^j), \\
    \mathcal{F}(\bm{\theta}) &=\text{log } P_{\bm{\theta}}(\mathbf{O}|\mathbf{W}^{\text{ref}})P(\mathbf{W}^{\text{ref}}) \notag\\
    &= \text{log } \sum_i P_{\bm{\theta}}(\bm{y}^{i,{ref}}_{1:T},\mathbf{O}|\mathbf{W}^{\text{ref}})P(\mathbf{W}^{\text{ref}}).
\end{align}
with $P_{\bm{\theta}}$ being a function of model parameters $\bm{\theta}$. By minimising the difference between the two objective functions, optimising w.r.t the MMI loss not only maximises the probability of $\mathbf{W}^{\text{ref}}$, but also minimises the probability  that of every competing hypothesis $\mathbf{W}^j$. This makes MMI a discriminatory loss.



\section{FST Lattice Based MMI Training}
\label{sec:fst}
The computation of $\mathcal{G}(\bm{\theta})$ involves summing over all possible sentence level state sequences which makes the computation  expensive. To address the computational overhead, the standard  FST lattice based approach employs a two step procedure: first a recognition pass on each training utterance is performed to collect only a subset of hypotheses that are most likely. The output of this process is then passed through a   determinization algorithm \cite{wfstlattice} using a special semiring that only preserves the best state alignment for each word-sequence. The resultant alignments are stored in a special FST called a lattice. The standard approach is to generate these lattices once using a good initialised model, which will be referred to as  the CE model going forward, and then proceed to use these lattices repeatedly at every iteration of discriminative sequence training. Although computing $\mathcal{F}(\bm{\theta})$ is computationally less intractable, the standard recipe in FST lattice based MMI training involves sampling the Viterbi alignment once using the CE model and using it as the only target alignment \cite{vesely}. For a given $\mathbf{O}$, these modifications lead to the  following proxy function  $\hat{\mathcal{L}}_{{MMI}}(\bm{\theta}) = \hat{\mathcal{G}}(\bm{\theta})  -  \hat{\mathcal{F}}(\bm{\theta})$ being minimised where
\begin{align}
\label{eq:KMMIloss}
     \hat{\mathcal{G}}(\bm{\theta}) &= \text {log} \sum_{j}^K P_{\bm{\theta}} \bigg( \hat{\bm{y}}^{\mathbf{W}^j}_{1:T}(\bm{\theta}_{\text{CE}}) ,\mathbf{O}|\mathbf{W}^j \bigg) P(\mathbf{W}^j), \\
    \hat{\mathcal{F}}(\bm{\theta}) &= \text {log } P_{\bm{\theta}} \bigg( \hat{\bm{y}}^{\mathbf{W}^{\text{ref}}}_{1:T}(\bm{\theta}_{\text{CE}}) ,\mathbf{O}|\mathbf{W}^{\text{ref}} \bigg).
\end{align} 
Here $\bm{\theta}_{\text{CE}}$ denotes the model parameters associated with CE model  and $\hat{\bm{y}}^{\mathbf{W}^j}_{1:T}(\bm{\theta}) =  \argmax_{\bm{y}^{i,j}_{1:T}} P_{\bm{\theta}}(\bm{y}^{i,j}_{1:T},\mathbf{O}| \mathbf{W}^j) P(\mathbf{W}^j)$.
FST lattice-based MMI training thus employs { an alignment-level} discriminative criterion where  at each iteration of training, the model is updated to reduce the {confusion} between a CE model chosen numerator alignment and the CE model chosen Viterbi alignments from  a set of competing hypothesis. 

Using $\hat{\mathcal{L}}_{{MMI}}(\bm{\theta}) $ as a proxy to the true MMI loss has been shown to lead to consistent Word Error Rate reduction (WERR) from discriminative sequence training \cite{vesely,li2021frame}. Although the procedure has been found to be experimentally effective, it is not clear why discriminating between a set of  pre-selected alignments correlates with  a hypothesis level discriminatory loss. Furthermore, it is biased to assume that a CE model chosen numerator alignment is the only feasible alignment for a given hypothesis. An utterance can have  multiple correct  alignments in its sentence level  $\mathbf{W}^{\text{ref}}$ HMM.   

The lower bound  in eqn.(\ref{lower_bound}) of the proof in Sec. \ref{den_proof} sheds some light in the effectiveness of such an approximation. The inequality shows how using a fixed subset of numerator alignments  throughout training can still guarantee improvement w.r.t the MMI objective.  This is verified in  Sec. \ref{sec:num} where  an empirical investigation on  the necessity in using a fixed pre-selected  numerator state alignment  is conducted. The effect of dynamically sampling a numerator state alignment from a CE model generated numerator lattice using the current model parameter update  is explored. In this work, two approaches to sampling have been considered.  The first one employs  the Viterbi algorithm \cite{Viterbi} to select  the best path at every iteration of training. The second approach employs ancestral sampling \cite{MattShannon} to sample paths from the lattice. In this latter approach samples are drawn efficiently using the backward filtering forward sampling algorithm \cite{loeliger2009estimating}. The algorithm performs   local normalisation of the weight leaving each state by re-weighting the weight with  the associated  probability scores  from the backward algorithm \cite{Baum1967}. Samples from the re-weighted FST can then be drawn using simple ancestral sampling.
\section{On the Fly  Lattice Determinization}
\label{sec:latdet}

This work also proposes using the following candidate function as a proxy to $\mathcal{G}(\bm{\theta})$:  
 \begin{align}
  \overline{ \mathcal{G}(\bm{\theta})} &= \text {log} \sum_{j}^K P_{\bm{\theta}} \bigg( \hat{\bm{y}}^{\mathbf{W}^j}_{1:T}(\bm{\theta}) ,\mathbf{O}|\mathbf{W}^j \bigg) P(\mathbf{W}^j).
\end{align}
The loss is computed by using the latest model update to select the Viterbi path associated with each hypothesis captured by the initial recognition pass using the CE model. The motivation behind this approach is that at each iteration of training, the model parameters will be updated to reduce the confusion between a chosen numerator alignment and  the alignment associated with  each competing  hypothesis that yields the greatest confusion. From an implementation point of view, this can be achieved by skipping the determinization process \cite{wfstlattice}   during the initial lattice generation and performing it on the fly during training. \emph{Theorem} \ref{den_proof} shows how such a modification in combination with  $\hat{\mathcal{F}}(\bm{\theta})$  guarantees discrimination over the subset of competing hypotheses captured in the lattice  w.r.t the MAP decision rule under certain assumptions. The efficacy of this proposed modified objective is investigated in Section \ref{sec:result}.

\section{Theorem }
\label{sec:theroem}

If the mapping $f: \{\mathbf {y}^i_{1:T} \}_i  \rightarrow \{\mathbf{W}^j\}_j $ is well defined then dynamically sampling the Viterbi paths of each competing hypothesis using the current model update in conjunction with using a fixed numerator alignment  guarantees discrimination w.r.t maximum a posteriori (MAP)  decision rule.

\label{den_proof}

\subsection{Discrete measure on the space of $\mathcal {Y}_{1:T}$}
  \label{discrete_measure}
 The mathematical proof presented relies on the concept of a measurable space and an associated measure \cite{axler2020measure}.
 Let  $\mathcal{X}$  denote the set of all state alignments $\{\mathbf {y}^i_{1:T} \}_i $  that are present in sentence level  hybrid HMM-DNN models associated with a given  $ \mathcal{O}$ with  $\mathcal{A}$ being the powerset of  $\mathcal{X}$.  Given that each state alignment corresponds to a hidden state sequence in some sentence level  HMM,  one can define as a   positive   real-valued function    $ {m} : (\mathcal{X}, \mathbb{R}^D) \rightarrow [0,\infty)$ as follows:
\begin{align}
{m}(\mathcal{Y}_{1:T},\bm{\theta}) &=  \sum_j \mathbf{1}_{\mathbf{W}^j}(\mathcal{Y}_{1:T}) P_{\bm{\theta}}(\mathcal{Y}_{1:T},\mathbf{W}^j|\mathcal{O}),
\end{align}
     where $ \mathcal {Y}_{1:T} \in \{\mathbf {y}^i_{1:T} \}_i$, $\bm{\theta} \in  \mathbb{R}^D $ denotes the $D$ dimensional  parameter vector associated with the model and $\mathbf{1}_{\mathbf{W}^j}$ is the indicator function which  equates to  1 if  $ {Y}_{1:T}$ belongs to the sentence level HMM of $\mathbf{W}^j$. To avoid notational clutter, the conditioning on  $\mathcal{O}$ is not explicitly stated in the definition of ${m}(\mathcal{Y}_{1:T},\bm{\theta})$.
    
    For any  vector $\bm{\theta}$, it  is easy to see that
 \begin{align}
    \sum_i {m}(\bm{y}^i_{1:T},\bm{\theta}) &\equiv 1.
 \end{align}
 Hence, one can define a discrete probability measure $ \mu_{\bm{\theta}} : \mathcal{A} \rightarrow [0,\infty] $ can now be defined as follows:
 \begin{align}
     \mu_{\bm{\theta}}(\mathbb{A}) = \sum_{i} {m}(\bm{y}^i_{1:T} |\bm{\theta}) \mathbf{1}_{\mathbb{A}} (\bm{y}^i_{1:T}) \mbox {  } \forall \mathbb{A} \in \mathcal{A}, \label{discrete_measure_}
 \end{align}

  where $\mathbf{1}_{\mathbb{A}} $ is the indicator function. It is easy to see that $\mu_{\bm{\theta}}(\mathcal{X} ) = 1$.

\subsection{Proof}
 
 Let $\mathbb{A}$ denote the set of all state alignments $\{\bm{y}^{i,{ref}}_{1:T}\}_i$, that belong to the sentence level  HMM model of  $\mathbf{W}^{{ref}}$,  and  let  $\bigcup_j \mathbb{B}_j$ be a union of measurable sets with each  $\mathbb{B}_j = \{\bm{y}^{i,j}_{1:T}\}_i$, the set of state alignments belonging to the constructed sentence level  HMM of $\mathbf{W}^j$. Thus, by construction
\[ \mathcal{X} = \bigg( \bigcup_j \mathbb{B}_j \bigg) \cup \mathbb{A}. \]
 Now let $\hat{\mathbb{A}} \subset \mathbb{A}$ be a measurable subset consisting of a single reference alignment $\{\bm{y}^{k,{ref}}_{1:T}\}$ and  $\hat{\mathbb{B}}_j \subset \mathbb{B}_j$ for each j such that
 \begin{align}
       \hat{\mathbb{B}}_j  &= \argmax_{\bm{y}^{i,j}_{1:T} \in  \mathbb{B}_j }   {m}(\bm{y}^{i,j}_{1:T},\bm{\theta}), \\
       &= \argmax_{\bm{y}^{i,j}_{1:T} \in  \mathbb{B}_j }  P_{\bm{\theta}}(\bm{y}^{i,j}_{1:T},\mathbf{W}^j|\mathcal{O}),  \\
       &= \argmax_{\bm{y}^{i,j}_{1:T} \in  \mathbb{B}_j } P_{\bm{\theta}}(\bm{y}^{i,j}_{1:T},\mathcal{O}|\mathbf{W}^j) P(\mathbf{W}^{{j}}).  \label{ eqn:vitalig}
\end{align}
 where  eqn (\ref{ eqn:vitalig} ) is a consequence of $P(\mathcal{O}) $ being a constant denominator term for all alignments.  As  $\mathcal{X}$ is a discrete  finite set,
 \begin{align}
      \mu_{\bm{\theta}}(\mathbb{B}_j) \leq \|\mathbb{B}_j \| \mbox {   }  \mu_{\bm{\theta}}   (\hat{\mathbb{B}}_j)   \label{discriminate_inequality}
 \end{align}
holds for all $j$ where $\mu_{\bm{\theta}}(\mathbb{B}_j)  = \sum_i {m}(\bm{y}^{i,{j}}_{1:T}|\bm{\theta})  = P_{\bm{\theta}}(\mathbf{W}^{{j}} |\mathcal{O})$.
 Similarly by  the monotonic property of the measure $ \mu_{\bm{\theta}}$, 
 \begin{align}
 \mu_{\bm{\theta}}(\hat{\mathbb{A}}) \leq  \mu_{\bm{\theta}}(\mathbb{A}), \label{lower_bound} 
 \end{align}
 where  $\mu_{\bm{\theta}}(\mathbb{A})  = \sum_i {m}(\bm{y}^{i,{ref}}_{1:T}|\bm{\theta})  = P_{\bm{\theta}}(\mathbf{W}^{{ref}} |\mathcal{O})$. 
 Keeping the set $\hat{\mathbb{A}}$ \textbf{fixed}, the measure $\mu_{\bm{\theta}}(\hat{\mathbb{A}}) $ is now a function of $\bm{\theta}  \in  \mathbb{R}^D $.  Eqn(\ref{lower_bound}) is analogous to the EM Algorithm where updating an auxiliary function iteratively improves $P_{\bm{\theta}}(\mathbf{W}^{{ref}} ,\mathcal{O})$. 
 
Consider the scenario where the set  $\hat{\mathbb{A}}$ is dynamically constructed by sampling a different alignment $\bm{y}^{j,{ref}}_{1:T}$ at each iteration of training. Although $m(\bm{y}^{j,{ref}}_{1:T}) \leq \mu_{\bm{\theta}}(\mathbb{A})$, it is not guaranteed that adapting the model parameters to increase the current sampled alignment will improve $\mu_{\bm{\theta}}(\mathbb{A})$ as increasing $m(\bm{y}^{j,{ref}}_{1:T}|\bm{\theta})$ may lead the probability of the previously sampled $\bm{y}^{k,{ref}}_{1:T}$ to decrease. Hence, keeping the numerator alignment fixed in essential to iteratively improve $P_{\bm{\theta}}(\mathbf{W}^{{ref}} ,\mathcal{O})$.



 Using a  sub collection of these constructed sets  $\{ \hat{\mathbb{B}}_j\}_{j=1} ^K$ and $\hat{\mathbb{A}}$,  a smaller measurable space $(\hat{\mathcal{X}},\hat{\mathcal{A}})$ can be constructed  with
$$ \hat{\mathcal{X}} = \bigg( \bigcup_{i=1} ^K \hat{\mathbb{B}_i} \bigg) \cup \hat{\mathbb{A}}, $$
and $\hat{\mathcal{A}}$ being the power set of this set. Using  ${\mu}_{\bm{\theta}}$, the following   probability measure $\hat{\mu}_{\bm{\theta}}$ can now be defined on this space:
\begin{align}
    \hat{\mu}_{\bm{\theta}} (C) =  \frac{ \mu_{\bm{\theta}} (C)} {  \sum_{j=1} ^K  \mu_{\bm{\theta}}( \hat{\mathbb{B}}_j)  } \mbox {for all } C \in \hat{\mathcal{A}}.
\end{align}
Under such a construction,
\begin{align}
     \hat{\mu}_{\bm{\theta}} (\hat{\mathbb{A}} ) & =   \frac{  \mu_{\bm{\theta}} (\hat{\mathbb{A}} )} {  \sum_{j=1} ^K  \mu_{\bm{\theta}}( \hat{\mathbb{B}}_j)  } \\
     &= \frac{P_{\bm{\theta}}(\mathcal{Y}_{1:T} = \bm{y}^{k,{ref}}_{1:T},\mathcal{O}|\mathbf{W}^{{ref}}) P(\mathbf{W}^{{ref}})}{\sum_{j=1}^K P_{\bm{\theta}}(\mathcal{Y}_{1:T} = \hat{\bm{y}}^{j}_{1:T},\mathcal{O}|\mathbf{W}^{{j}}) P(\mathbf{W}^{{j}})},
\end{align}
where 
$ \hat{\bm{y}}^{j}_{1:T} =\argmax_i P_{\bm{\theta}}(\mathcal{Y}_{1:T} = \hat{\bm{y}}^{j,i}_{1:T},\mathcal{O}|\mathbf{W}^{{j}}) P(\mathbf{W}^{{j}}) $. The negative $\rm{log}$ of $\hat{\mu}_{\bm{\theta}} (\hat{\mathbb{A}} )$ corresponds to the proposed FST lattice-based MMI objective with the Viterbi path  of each competing hypothesis chosen by the current model update. Since logarithmic function is monotonic, minimising  $ -\text{log} \big(\hat{\mu}_{\bm{\theta}} (\hat{\mathbb{A}} ) \big)$ is equivalent to maximising $\hat{\mu}_{\bm{\theta}} (\hat{\mathbb{A}} )$. For any parameter vector  $\bm{\theta} \in  \mathbb{R}^D $, the associated measure $\mu_{\bm{\theta}}$ will all satisfy the inequalities (\ref{discriminate_inequality}) and (\ref{lower_bound}).  Therefore maximising $\hat{\mu}_{\bm{\theta}} (\hat{\mathbb{A}}) $ can be seen to maximise the probability of $\mathbf{W}^{{ref}}$ while simultaneously decreasing the probability of the competing hypothesis captured in the lattice as a consequence of  lower bound (\ref{discriminate_inequality}). This ensures discrimination  w.r.t MAP decision rule.

\section{Experimental Setup}
\label{sec:setup}

The experiments relevant to this work were conducted on an  internal anonymized Mandarin and British English dataset. For the  Mandarin ASR experiment, 18K hours of training data was used from which roughly 10 hrs of training data was uniformly sampled  to form the validation set.  For the British English ASR experiment, systems were trained using a subset of 2.8K hours from the British English training data from which 15 hours of uniformly sampled data was used as the validation set. To estimate the generalisation performance of the candidate models, decoding of the resultant models  was performed on  language specific test sets using a pre-trained neural language model. In the Mandarin ASR task, 4 independent test sets: two comprising  of assistant task data and two consisting of dictation data were used. While for  the  British English ASR task, an  assistant task data set and  dictation data test set  were used respectively. Each of these sets composed roughly of 36 hrs of data.

The efficacy of the proposed  modification to the FST lattice based MMI  denominator loss computation and  the investigation on  the effect of dynamically sampling a numerator alignment is shown on  training a  50-layer self-normalizing deep CNN (SNDCNN) model \cite{RFDNN} \footnote{ a Resnet-50 model with the residual connections removed and the ReLU activation function replaced by the SeLU activation.}. The  input to the model was produced by splicing together 80 dimensional log-Mel filter bank (FBK) features  using a context window
of 41 frames. For all experiments, the input features were  mean and variance normalised w.r.t the training data. For the Mandarin ASR experiment, the filter bank features were augmented with pitch information \cite{6854049} and frame level spec-augmentation \cite{li2021frame} was employed.

Prior to sequence training, the models were {initialised} with  frame-level cross entropy training to serve as  the CE models. These models were used to create the denominator lattices.  Determinized lattices were used to get baseline WERs  while to train models with the proposed modification,  non-determinized lattices were generated.  To investigate the effect of dynamically sampling a numerator alignment,  numerator lattices were created under the same conditions used to generate the  non-determinized denominator lattices.   As there is an obvious mismatch between the various training criteria explored  with the WER, over-fitting to the training criterion can occur \cite{haider2018}. To track how generalisation improvements w.r.t the training criterion correlates with WER reduction and to perform  model selection, additional decoding  was performed using a language specific separate held out development set after each epoch. 

 An epoch size of {500 hours} was used in the British English ASR experiment . Models were trained in a decentralized distributed setting \cite{9585552} using the  ADAM optimiser \cite{kingma2014adam} equipped  with the new bob  learning rate scheduler.  For the much larger Mandarin ASR experiment, an epoch size  of  200 hrs was used. To ensure fast and efficient training, the Block Model Update filtering (BMSGD) \cite{chen2016scalable} algorithm was used to train the models in a centralized distributed  setting.

\section{Summary of Results}
\subsection{Efficacy Of Using Fixed Numerator Alignment \label{sec:num}}
The  effect of dynamically sampling the numerator alignment during MMI training was evaluated on the British English ASR task. Table \ref{tab:num sampled} summarizes the results where `baseline' correponds to the standard MMI recipe of using a fixed CE model  chosen numerator alignment.  The use of a fixed alignment can be seen to yield the greatest WERR from FST lattice based MMI training which corresponds to a relative improvement of 15\%  WERR on  both the dictation and assistant task test sets.  Over the two sampling based approaches, the fixed alignment approach achieves a relative WERR of 9.5-10\% on the dictation task and 2.8\% on the assistant task. 
 Inequality (\ref{lower_bound}) provides some insight into this behaviour: maximising the probability of fixed set of numerator alignments acts as a lower bound to the MMI objective's numerator component.

 \begin{table}
     \centering
     
     \resizebox{6.5cm}{!}{%
     \begin{tabular}{|c|c|c|}
     \hline
     Model   & Assistant & Dictation \\
     \hline
    CE Model  & 7.26 & 7.82 \\
     \hline
     \multicolumn{3}{c}{ MMI Training} \\
     \hline
    Numerator alignment selection  & Assistant & Dictation \\
    \hline
        Baseline      &   \textbf{6.16}            &       \textbf{ 6.57} \\
      \hline
        Ancestral sampling  & 6.34          &          7.33 \\
      \hline
        Viterbi   &  6.33       &             7.26 \\
       
        \hline
     \end{tabular}
     }
     \caption{WERs on the British English test sets with different choices of numerator state alignment used during MMI training.}
      \label{tab:num sampled}
 \end{table}

\subsection{{On The Fly Denominator Lattice determinization} \label{sec:result}}
 
 The standard FST lattice based approach employs a two step procedure:  first a recognition pass on each training utterance is performed using the CE model to collect only a subset of hypotheses that are most likely.  As mentioned in Sec. \ref{sec:fst}, the output of this process is then passed through a determinization algorithm that only preserves the best state alignment for each word-sequence.  In the proposed modification to MMI training, step 2 is deferred to the actual training stage. The current model update is used to select the best state alignment for each competing word-sequence during training. As a consequence, the lattices stored will maintain multiple alignments per hypothesis leading to increased lattice size. For the datasets used in this experiment, the storage overhead was found to increase by a factor of 5.  The issue can be resolved by generating the lattices on the fly but this comes at the expense of increased training time. The proposed approach lies in between the lattice-based and on the fly lattice generation approach to MMI training. The non-necessity to do a recognition pass  makes it less computationally expensive than creating the lattices on the fly.


 \begin{table}
 \resizebox{8.9cm}{!}{%
    \begin{tabular}{|c|c|c|c|c|}
        \hline
        Approach &  Assistant  & Assistant & Dictation & Dictation \\
         & task 1 & task 2 & task 1 & task 2\\
         \hline
         CE Model  & 8.55 &	8.53 &	7.50&	7.51\\
         
         \hline
          \multicolumn{5}{c}{ MMI Training} \\
        \hline
           Baseline &  7.37 &	7.44 &	6.57 &	6.49\\
            \hline
            \hline
            MMI+ on the fly & \textbf{7.20} &	\textbf{7.21} &	\textbf{6.29} &	\textbf{6.27} \\
           determinization   &  &   & &\\ 
            \hline
            \hline
    \end{tabular}
    }
    \centering
    \caption{WERs with variants of FST lattice based MMI loss on the Mandarin test sets \label{proposed_big}}
\end{table}

 

 Table  \ref{proposed_big} compares the efficacy of the resultant Mandarin acoustic model trained with  the proposed approach against an equivalent  CE initialized model trained with  the standard FST lattice based recipe on the 4 test sets. Sequence training using the standard approach  can be seen to lead to a relative improvement of 12.7-13.8\% on the  assistant test sets and 12.4-13.6\% WERR on the dictation test sets.  Over the standard approach, the proposed approach achieves a further relative improvement of 2.3-3.1\% on the assistant tasks and 3.4-4.6\% WERR on the dictation  test sets. The improvements seen accompanied by the  mathematical proof in  Sec \ref{sec:theroem} presents a strong argument in the efficacy of the proposed modification to the FST lattice based MMI loss computation.

 Table \ref{proposed_small} compares the efficacy of models trained on the British English dataset. Over the baseline approach, the proposed approach achieves a relative improvement of 2.3\% on the assistant task  set and 1.0\% on the dictation task test set. The table also shows the generalisation performance of a training setup that is analogous with the on-the-fly FST lattice  generation based training with the exception that the initial recognition pass is performed with the CE model. From comparing the results with  Table \ref{tab:num sampled}, determinizing the denominator lattice on the fly  on a setup where the numerator path is dynamically sampled using the Viterbi algorithm leads to an increased WER improvement of 5\% and 2.3\%   on the dictation and assistant task test set respectively. Thus, discriminating between a fixed numerator alignment and  the viterbi alignments associated with  each competing  hypothesis   chosen by the current model update  can be seen to be the most effective in achieving better  WERRs from FST lattice based MMI training.

\begin{table}
\resizebox{6.5cm}{!}{%
    \begin{tabular}{|c|c|c|}
        \hline
        Approach &  Assistant & Dictation \\
        \hline
           Baseline & 6.16 & 6.57\\
            \hline
            Num path: Viterbi  & &\\
           +  on the fly  determinization & 6.18 &  6.90\\ 
            \hline
            MMI + on the fly  & & \\
           determinization   & \textbf{6.02} &  \textbf{6.53}\\ 
            \hline
            \hline
    \end{tabular}
    }
    \centering
    \caption{WERs with variants of FST lattice based MMI loss  on the British English dataset. \label{proposed_small}}
\end{table}

 \section{Conclusion}
This paper has highlighted the  various implicit modifications  introduced to the MMI objective when models are trained in an FST lattice based framework. To this end, one of the major goals of this work has been to bring  forward  these details to a larger audience in the ASR community and allow more people to work on the particular challenges of FST based MMI training.  The paper  in particular investigated the necessity to maintain a pre-selected  numerator alignment and addressed the importance of on the fly  FST denominator lattice determinization. The effectiveness of employing on the fly FST lattice determinization is  mathematically supported  to guarantee  discrimination at the hypothesis level  and is  empirically shown on training a SNDCNN model  on an 18K Mandarin dataset and 2.8K British English dataset respectively. On the Mandarin assistant and dictation test sets, the proposed approached achieved between  2.3-3.1\% WERR on the assistant test sets and between 3.4-4.6\% WERR on the dictation  test sets over the standard FST lattice MMI training.

\bibliographystyle{IEEEbib}
\bibliography{strings}

\begin{thebibliography}{10}

\bibitem{bahl1986maximum}
L.~Bahl, P.~Brown, P.V. de~Souza, and R.~Mercer,
\newblock ``Maximum mutual information estimation of hidden {M}arkov model
  parameters for speech recognition,''
\newblock in {\em Proceedings of the 11th IEEE International Conference on
  Acoustics, Speech and Signal Processing (ICASSP)}, 1986, pp. 231--234.

\bibitem{Baum1967}
L.E. Baum and J.A. Eagon,
\newblock ``An inequality with applications to statistical estimation for
  probabilistic functions of {M}arkov processes and to a model for ecology,''
\newblock {\em Bulletin of the American Mathematical Society}, vol. 73(3), pp.
  360--363, 1967.

\bibitem{Xiong:2016a}
W.~Xiong, J.~Droppo, X.~Huang, F.~Seide, M.~Seltzer, A.~Stolcke, D~Yu, and
  G.~Zweig,
\newblock ``The microsoft 2016 conversational speech recognition system,''
\newblock in {\em Proceedings of the 41st IEEE International Conference on
  Acoustics, Speech and Signal Processing (ICASSP)}, 2016, pp. 5255--5259.

\bibitem{Saon:2016a}
G.~Saon, T.~Sercu, S.~Rennie, and H.-K.J. Kuo,
\newblock ``The {IBM} 2016 {E}nglish conversational telephone speech
  recognition system,''
\newblock in {\em Proceedings of the 17th Conference of the International
  Speech Communication Association (Interspeech)}, 2016, pp. 7--11.

\bibitem{MMI2}
P.C. Woodland and D.~Povey,
\newblock ``Large scale discriminative training of hidden {M}arkov models for
  speech recognition,''
\newblock {\em Computer Speech and Language}, vol. 16(1), pp. 25--47, 2002.

\bibitem{Povey2005}
Daniel Povey,
\newblock {\em Discriminative training for Large Vocabulary Speech
  Recognition},
\newblock Ph.D. thesis, University of Cambridge, 2005.

\bibitem{htk}
S.~Young, G.~Evermann, M.~Gales, T.~Hain, D.~Kershaw, X.~Liu, G.~Moore,
  J.~Odell, D.~Ollason, D.~Povey, A.~Ragni, V.~Valtchev, P.~Woodland, and
  C.~Zhang,
\newblock {\em The HTK Book (for HTK version 3.5)},
\newblock Cambridge University Engineering Department, 2015.

\bibitem{kaldi}
D.~Povey, A.~Ghoshal, G.~Boulianne, L.~Burget, O.~Glembek, N.~Goel,
  M.~Hannemann, P.~Motl\'{i}\v{c}ek, Y.~Qian, P.~Schwarz, J.~Silovsk\'{y},
  G.~Stemmer, and K.~Vesel\'{y},
\newblock ``The {K}aldi speech recognition toolkit,''
\newblock in {\em Proceedings of the 8th IEEE Workshop on Automatic Speech
  Recognition and Understanding (ASRU)}, 2011, pp. 1--6.

\bibitem{mohri2002weighted}
Mehryar Mohri, Fernando Pereira, and Michael Riley,
\newblock ``Weighted finite-state transducers in speech recognition,''
\newblock {\em Computer Speech \& Language}, vol. 16, no. 1, pp. 69--88, 2002.

\bibitem{wfstlattice}
Daniel Povey, Mirko Hannemann, Gilles Boulianne, Lukáš Burget, Arnab Ghoshal,
  Miloš Janda, Martin Karafiát, Stefan Kombrink, Petr Motlíček, Yanmin
  Qian, Korbinian Riedhammer, Karel Veselý, and Ngoc~Thang Vu,
\newblock ``Generating exact lattices in the {WFST} framework,''
\newblock in {\em 2012 IEEE International Conference on Acoustics, Speech and
  Signal Processing (ICASSP)}, 2012, pp. 4213--4216.

\bibitem{vesely}
Karel Vesel{\`y}, Arnab Ghoshal, Luk{\'a}s Burget, and Daniel Povey,
\newblock ``Sequence-discriminative training of deep neural networks,''
\newblock in {\em Proceedings of the 14th Conference of the International
  Speech Communication Association (Interspeech)}, 2013, pp. 2345--2349.

\bibitem{li2021frame}
Xinwei Li, Yuanyuan Zhang, Xiaodan Zhuang, and Daben Liu,
\newblock ``Frame-level specaugment for deep convolutional neural networks in
  hybrid {ASR} systems,''
\newblock in {\em IEEE Spoken Language Technology Workshop (SLT)}, 2021, pp.
  209--214.

\bibitem{Viterbi}
Andrew Viterbi,
\newblock ``Error bounds for convolutional codes and an asymptotically optimum
  decoding algorithm,''
\newblock {\em IEEE transactions on Information Theory}, vol. 13, no. 2, pp.
  260--269, 1967.

\bibitem{MattShannon}
Matt Shannon,
\newblock ``Optimizing expected word error rate via sampling for speech
  recognition,''
\newblock in {\em Proceedings of the 18th Conference of the International
  Speech Communication Association (Interspeech)}, 2017, pp. 3537--3541.

\bibitem{loeliger2009estimating}
Hans-Andrea Loeliger and Mehdi Molkaraie,
\newblock ``Estimating the partition function of 2-d fields and the capacity of
  constrained noiseless 2-d channels using tree-based {Gibbs} sampling,''
\newblock in {\em Proceedings of IEEE Information Theory Workshop}, 2009, pp.
  228--232.

\bibitem{axler2020measure}
Sheldon Axler,
\newblock {\em Measure, integration \& real analysis},
\newblock Springer Nature, 2020.

\bibitem{RFDNN}
Zhen Huang, Tim Ng, Leo Liu, Henry Mason, Xiaodan Zhuang, and Daben Liu,
\newblock ``{SNDCNN:} self-normalizing deep {CNN}s with scaled exponential
  linear units for speech recognition,''
\newblock in {\em Proceedings of International Conference on Acoustics, Speech
  and Signal Processing (ICASSP)}, 2020, pp. 6854--6858.

\bibitem{6854049}
Pegah Ghahremani, Bagher Baba~Ali, Daniel Povey, Korbinian Riedhammer, Jan
  Trmal, and Sanjeev Khudanpur,
\newblock ``A pitch extraction algorithm tuned for automatic speech
  recognition,''
\newblock in {\em 2014 IEEE International Conference on Acoustics, Speech and
  Signal Processing (ICASSP)}, 2014, pp. 2494--2498.

\bibitem{haider2018}
Adnan Haider and P.C. Woodland,
\newblock ``Combining natural gradient with {H}essian free methods for sequence
  training,''
\newblock in {\em Proceedings of the 19th Conference of the International
  Speech Communication Association (InterSpeech)}, 2018, pp. 2918--2922.

\bibitem{9585552}
Xiaodong Cui, Wei Zhang, Abdullah Kayi, Mingrui Liu, Ulrich Finkler, Brian
  Kingsbury, George Saon, and David Kung,
\newblock ``Asynchronous decentralized distributed training of acoustic
  models,''
\newblock {\em IEEE/ACM Transactions on Audio, Speech, and Language
  Processing}, vol. 29, pp. 3565--3576, 2021.

\bibitem{kingma2014adam}
Diederik~P Kingma and Jimmy~Lei Ba,
\newblock ``{Adam}: {A} method for stochastic optimization,''
\newblock in {\em Proceedings of the 3rd International Conference on Learning
  Representations (ICLR)}, 2015, pp. 1--13.

\bibitem{chen2016scalable}
Kai Chen and Qiang Huo,
\newblock ``Scalable training of deep learning machines by incremental block
  training with intra-block parallel optimization and blockwise model-update
  filtering,''
\newblock in {\em Proceedings of the 41st IEEE International Conference on
  Acoustics, Speech and Signal Processing (ICASSP)}, 2016, pp. 5880--5884.

\end{thebibliography}

\end{document}